\documentclass[conference]{IEEEtran}
\IEEEoverridecommandlockouts
\usepackage[left=0.75in, right=0.75in, top=0.75in, bottom=0.75in]{geometry}
\usepackage{cite}
\usepackage{gensymb}
\usepackage{comment}
\usepackage{amsmath,amssymb,amsfonts}
\usepackage{algorithmic}
\usepackage{graphicx}
\usepackage{textcomp}
\usepackage{xcolor}
\usepackage{siunitx}
\usepackage{url}
\usepackage{svg}
\usepackage[hidelinks]{hyperref}

\usepackage{subcaption}

\def\BibTeX{{\rm B\kern-.05em{\sc i\kern-.025em b}\kern-.08em
    T\kern-.1667em\lower.7ex\hbox{E}\kern-.125emX}}
\begin{document}

\bstctlcite{BSTcontrol}

\title{\vspace{0.25in}Path-conditioned Reinforcement Learning-based Local Planning for Long-Range Navigation
}

\newif\ifanonymous
\anonymousfalse 

\DeclareRobustCommand{\anontext}[2]{%
  \ifanonymous
    \textcolor{orange}{[#2]}%
  \else
    #1%
  \fi
}

\author{
\anontext{
Mateo Haro, Julia Richter, Fan Yang, Cesar Cadena, Marco Hutter
\thanks{All authors are with the Robotic Systems Lab (RSL), ETH Zürich, Zürich, Switzerland}%
\thanks{Corresponding author: Mateo Haro,  {\tt\small maharo@ethz.ch}.}%
}
{anonymous authors}
\thanks{Project website: \href{https://leggedrobotics.github.io/rl-path-following/}{leggedrobotics.github.io/rl-path-following}}%
}

\maketitle

\begin{abstract}
Long-range navigation is commonly addressed through hierarchical pipelines in which a global planner generates a path, decomposed into waypoints, and followed sequentially by a local planner. 
These systems are sensitive to global path quality, as inaccurate remote sensing data can result in locally infeasible waypoints, which degrade local execution. At the same time, the limited global context available to the local planner hinders long-range efficiency.
To address this issue, we propose a reinforcement learning–based local navigation policy that leverages path information as contextual guidance. 
The policy is conditioned on reference path observations and trained with a reward function mainly based on goal-reaching objectives, without any explicit path-following reward.
Through this implicit conditioning, the policy learns to opportunistically exploit path information while remaining robust to misleading or degraded guidance.
Experimental results show that the proposed approach significantly improves navigation efficiency when high-quality paths are available and maintains baseline-level performance when path observations are severely degraded or even non-existent. 
These properties make the method particularly well-suited for long-range navigation scenarios in which high-level plans are approximate and local execution must remain adaptive to uncertainty.
\end{abstract}
\vspace{0.2cm}

\section{Introduction}

Autonomous navigation is a fundamental capability of robotic systems and has become increasingly important as robots are deployed with higher levels of autonomy in complex, real-world environments. While planning and navigation have a long-standing history in robotics, the field has been profoundly reshaped in recent years by advances in deep learning and reinforcement learning (RL) \cite{10.1145/3727642,nguyen2025emergencedeepreinforcementlearning,anderson2018evaluationembodiednavigationagents}. Learning-based navigation systems offer improved robustness to uncertainty, partial observability, and sensor noise, and generalize well to dynamic and previously unseen environments \cite{fan2018crowdmove,shah2022viking,truong2024indoorsim,yang2025spatially}. These properties make RL particularly attractive for long-range navigation, where explicit maps may be unavailable or inaccurate. However, despite this progress, enabling RL agents to scale efficiently to long-horizon navigation tasks remains a central challenge, as pure goal-oriented exploration often leads to inefficient trajectories and poor utilization of available high-level information \cite{yang2025spatially,richter2026large}.

Most navigation systems are structured hierarchically, decomposing the task into a high-level global planner and a low-level local planner \cite{alma9936230413403606,mattamala2024autonomous,richter2026large}. The global planner is typically responsible for computing a path from the robot’s current location to the final goal, often based on coarse or incomplete map information, while the local planner executes short-range motions toward intermediate goals or waypoints. In many existing approaches, the global planner is typically assumed to provide an ideal and fully feasible path, and the local planner is often encouraged to follow this path as closely as possible \cite{hentschel2010autonomous,app12146874}. Other methods provide the local planner with only limited global context, such as a single waypoint \cite{mattamala2024autonomous,richter2026large}. As a consequence, the local policy must rely heavily on exploration and reactive behaviors, which can lead to inefficiencies, repeated traversal of dead ends, and suboptimal long-range performance. In both cases, the local planner lacks a global understanding of the intended trajectory, either due to over-reliance on perfect guidance or insufficient access to global information \cite{richter2026large}. While relatively few approaches provide local planners with rich global context without enforcing path adherence \cite{shah2022viking,truong2024indoorsim}, recent surveys such as \cite{nguyen2025emergencedeepreinforcementlearning} point to the integration of heuristic guidance into deep reinforcement learning (DRL) as a promising avenue for improving efficiency and long-range navigation performance.

\begin{figure}
    \centering
    \includegraphics[width=1\linewidth]{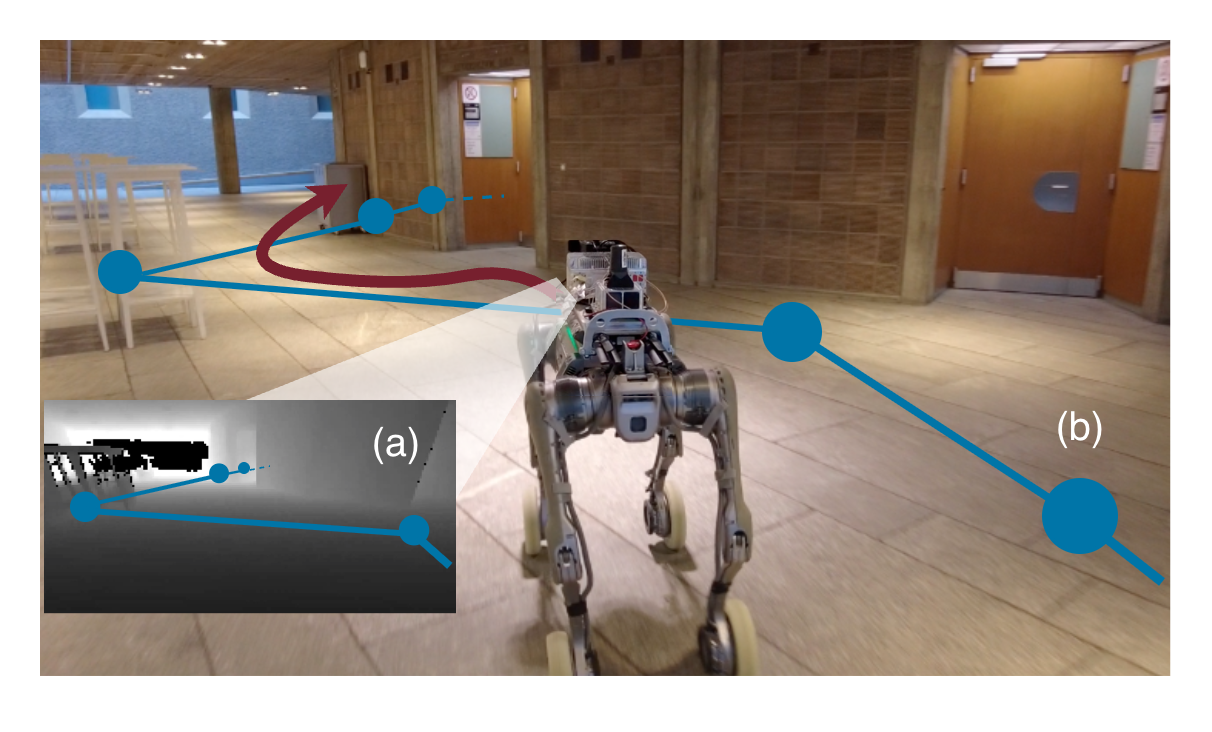}
    \caption{The proposed path-aware local planner utilizes a reference path (b) to provide contextual guidance, augmenting depth camera observations (a) and proprioceptive feedback.}
    \label{fig:intro_fig}
\end{figure}

In this work, we take a step in this direction by introducing a path-aware RL framework that incorporates coarse global path information as a form of heuristic guidance, without assuming this guidance is optimal or directly traversable. Rather than enforcing explicit path following, the policy learns to selectively exploit global path cues when they are informative and to disregard them when they are misleading, as illustrated in Fig.~\ref{fig:intro_fig}. We achieve this by extending a state-of-the-art RL-based navigation framework \cite{yang2025spatially} with a structured path representation and a training procedure that integrates the global path information. 
Concretely, our contributions are as follows:

\begin{enumerate}
    \item \textbf{Path-conditioned navigation architecture:}
    We introduce a path-aware policy architecture that learns a structured representation of the global path and fuses it with perceptual observations.
    \item \textbf{Path-conditioned training strategy:}
    We propose a training strategy that enables the policy to remain effective when path guidance is noisy or suboptimal.
    \item \textbf{Real-world deployment:} We validate the proposed approach through various simulation experiments and real-world deployment, assessing architectural choices, training procedures, and robustness to noisy and suboptimal path observations.
\end{enumerate}
The code is publicly released for the research community.

\section{Related Work}

Historically, search-based algorithms, such as A* \cite{Hart1968}, have often been used alongside sampling-based methods \cite{LaValle1998RapidlyexploringRT, kavraki2002probabilistic} to balance planning efficiency and exploration capabilities in discrete environments. 
The emergence of learning-based methods has further improved navigation capabilities by increasing adaptability to environmental uncertainty \cite{shah2022viking,Lee_2024,yang2025spatially}. 
Many recent approaches integrate learned local planners alongside classical, non-learning-based global planners, reducing reliance on hand-crafted rules and extensive hyperparameter tuning \cite{fan2018crowdmove,Lee_2024}. 
However, these systems still typically depend on explicit environment mapping, which comes with downsides as storage issues, dynamic obstacles, and a lack of map quality, as pointed out by \cite{10.1145/3727642}.

In contrast to a hierarchical approach, several works have explored end-to-end DRL approaches for autonomous navigation in unknown environments \cite{fan2018crowdmove,shi2019end,cimurs2021goal}. These methods aim to remove explicit planning and mapping stages by learning reactive policies that map sensory inputs directly to control actions. While such approaches demonstrate strong autonomy and adaptability, multiple studies report reduced navigation efficiency due to the lack of global map information, often resulting in longer and more circuitous trajectories \cite{shi2019end,cimurs2021goal}. 

Another line of research introduces plan- or path-conditioned reinforcement learning policies. These methods provide the agent with a reference trajectory and design reward structures that explicitly encourage adherence to the given path. The theory of plan-based shaping rewards \cite{schubert2021plan} formalizes this approach and has been successfully applied theoretically \cite{9205217, schubert2021learning} as well as to autonomous driving, where precise path tracking is critical \cite{merton2024deep,app12146874}. However, such formulations inherently constrain the agent’s behavior, limiting its ability to explore or deviate from the reference path even when cheaper or safer alternatives exist.

The work most closely related to ours is L2E (Learn to Execute) \cite{schubert2021learning}, which proposes a plan-conditioned policy capable of reaching a goal even when the provided plan is partially incorrect. While L2E allows deviations when a plan becomes infeasible, its reward structure explicitly enforces proximity to the given trajectory whenever it is valid. Moreover, the approach is demonstrated primarily on manipulation tasks, which differ significantly from navigation in large-scale environments. In contrast, our approach does not aim to strictly execute a provided plan. Instead, we leverage the plan as contextual information while allowing the agent to opportunistically deviate whenever more efficient alternatives are observed, even if the original path remains feasible.

Attention mechanisms have also been explored in local planning, but typically not to encode the reference path or goal itself \cite{pokle2019deep, shah2023vint}. FocusNav \cite{zhang2026focusnav} uses attention to predict its own collision-free waypoints, rather than to encode an externally provided path. Pokle et al. \cite{pokle2019deep} and ViNT \cite{shah2023vint} embed the global path or goal via an MLP or CNN rather than attention. None of these approaches uses attention to encode the path itself before fusing it at a later stage of the architecture. 


Overall, our work seeks to bridge pure goal-oriented exploration, DRL navigation, and path-conditioned control by combining the flexibility of learned policies with the guidance of a reference path, without explicitly enforcing path following, enabling a local planner that can be integrated into hierarchical navigation systems and intelligently exploit global planner path information.

\section{Problem Formulation}
We consider the navigation problem introduced in~\cite{yang2025spatially}, where an agent operates in an unknown 3D environment $E \subset \mathbb{R}^3$ and must reach a goal specified as a relative position $p_t \in \mathbb{R}^3$ to the agent's egocentric frame at time step $t$, using only a single front-facing depth camera as exteroceptive input. The task is defined as reaching the goal region
\begin{equation}
\mathcal{G} = \{ p_t \in \mathbb{R}^3 \mid \lVert p_t \rVert < \epsilon \}
\end{equation}
within a maximum time horizon $t \leq T_{\max}$. Because the agent receives egocentric observations $o_t$ that do not fully capture the underlying environment state $s_t$, the problem is formulated as a Partially Observable Markov Decision Process (POMDP) defined by the tuple $(\mathcal{S}, \mathcal{A}, \mathcal{T}, \mathcal{R}, \mathcal{O}, \mathcal{Z}, \gamma)$, where: $\mathcal{S}$ is the environment state space, $\mathcal{A}$ the agent action space, $\mathcal{T}(s,a,s')$ the transition probability function, $\mathcal{R}(s,a)$ the reward function, $\mathcal{O}$ the agent observation space, $\mathcal{Z}(s,a,o)$ the observation probability function and $\gamma \in [0,1]$ is the discount factor.


At each time step $t$, the agent receives an observation $o_t$ and selects an action $a_t$ according to a policy $\pi$. Since the state is not directly observable, the current observation is fused with the observation history $\mathcal{H}_{t-1} = \{o_1,\dots,o_{t-1}\}$ through a function $f$ that produces a state estimate: $\hat{s}_t = f(o_t, \mathcal{H}_{t-1})$.
The policy then computes the action $a_t = \pi(\hat{s}_t)$.
Both the state estimation function $f$ and the policy $\pi$ are jointly learned end-to-end via reinforcement learning, as described in the next section.

\section{Methodology}
\label{sec:methodology}

\subsection{Path Representation and Encoding}

Providing the full reference path allows the policy to reason over global structure, such as upcoming turns, large-scale deviations, or alternative routes, which cannot be captured by local look-ahead representations. Since our objective is to enable anticipatory behavior over long horizons, we encode the entire reference path rather than a truncated segment of arbitrary length, which would require task-specific tuning. Full-path observations offer a more general and robust representation, particularly in complex terrain and large-scale navigation scenarios.
At time step $t$, the observation is defined as
\begin{equation}
o_t \in \mathcal{O}, \quad 
o_t = \{ I_t,\; p_t,\; o_t^{prop},\; \tilde{P}_t \},
\end{equation}
where $I_t$ denotes the depth image, $p_t$ the relative goal position, $o_t^{prop}$ the proprioceptive observations, and $\tilde{P}_t$ the relative reference path representation. 
The construction of $\tilde{P}_t$ is described below. 
The reference path $P_t$ is defined as a fixed-length sequence of $N$ waypoints expressed in the robot's egocentric frame:
\begin{equation}
P_t = \{ p_i \}_{i=1}^{N}, 
\quad p_i \in \mathbb{R}^3.
\end{equation}
For each waypoint $p_i$, we compute its Euclidean distance to the robot:
\begin{equation}
d_i = \| p_i \|_2.
\end{equation}
We then construct an augmented waypoint representation $\tilde{p}_i \in \mathbb{R}^4$ defined as
\begin{equation}
\tilde{p}_i = 
\begin{bmatrix}
\displaystyle \frac{p_i}{\max (d_i, \epsilon)} \\
\displaystyle \frac{\log(1 + d_i)}{c}
\end{bmatrix},
\end{equation}
where $\frac{p_i}{\max (d_i, \epsilon)}$ encodes the normalized direction of the waypoint, with $\epsilon$ a small constant preventing division by zero, and $\log(1 + d_i)$ compresses large distance variations. The constant $c=\max _i(\log(1+ d_i))$ denotes a normalization factor ensuring bounded inputs for training stability. The final path representation $\tilde{P}_t$ encodes the waypoints' direction and log-distance to the current robot pose and is therefore
\begin{equation}
\tilde{P}_t = \{ \tilde{p}_i \}_{i=1}^{N}, 
\quad \tilde{p}_i \in \mathbb{R}^4.
\end{equation}
This relative encoding enables generalization across varying path lengths and geometries, while absolute distance information remains accessible through $p_t$. Note that $p_t$ is not constrained to coincide with the final waypoint $p_N$, as the reference path may not necessarily align with the goal position, for example due to the noise introduced during training, as described in the following section.

\begin{figure}
    \centering
    \includegraphics[width=1\linewidth]{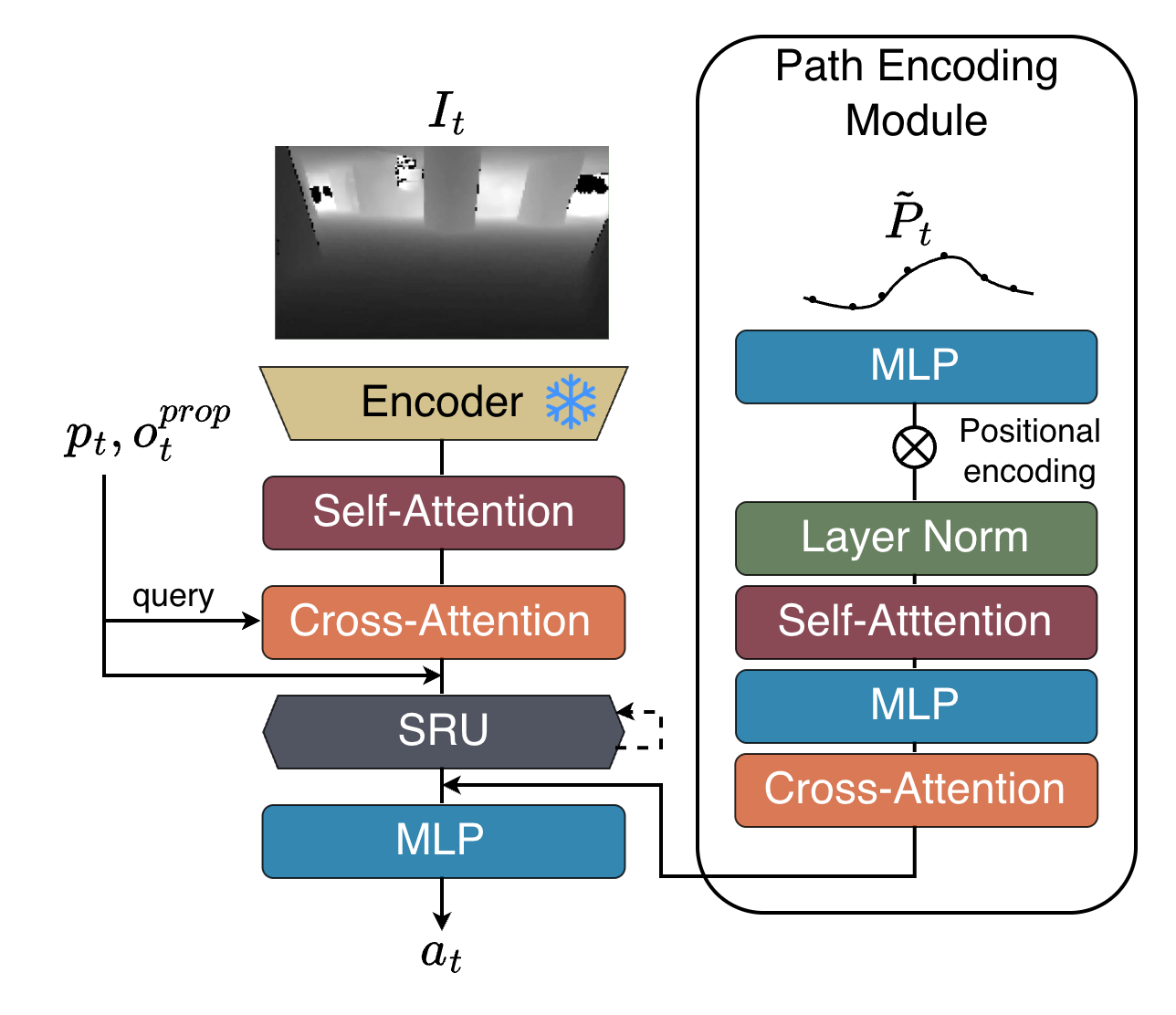}
    \caption{Overview of the proposed path-aware navigation architecture. The base navigation pipeline, including the pretrained depth image encoder, the spatially-enhanced recurrent unit, the fusion of relative goal position $p_t$ and proprioceptive observations $o_t^{prop}$ with visual features, follows \cite{yang2025spatially}. We extend this framework with the proposed Path Encoding module, enabling implicit conditioning on reference path information.}
    \label{fig:full_architecture}
\end{figure}

To encode the path, we employ a combination of self-attention and cross-attention mechanisms, as illustrated in Fig.~\ref{fig:full_architecture}. Self-attention enables interaction among waypoints and captures global path structure, producing a compact set of informative features. While self-attention alone could serve as a dimensionality-reduction mechanism, we also introduce cross-attention to allow the policy to selectively focus on relevant segments of the path. Specifically, the cross-attention query is implemented as a learnable embedding of fixed dimension, enabling the model to attend to path regions that are most informative for the current navigation context, such as sharp turns or critical deviations. The resulting path embedding is concatenated with the output of the spatially-enhanced recurrent unit (SRU), as illustrated in Fig.~\ref{fig:full_architecture}. In contrast, proprioceptive features are integrated upstream of the SRU and contribute to its hidden state. This design reflects the assumption that the reference path remains fixed during an episode and therefore does not require temporal integration within the recurrent memory.

\subsection{Training Path Generation}

Rather than training exclusively on optimal trajectories, we sample reference paths from a controlled distribution that includes feasible but suboptimal solutions. This prevents the policy from collapsing to strict path-following behavior while preserving informative global guidance. To further increase robustness, we intentionally perturb paths with noise, which can make them locally infeasible.

Optimal reference paths are generated using A* search on a graph constructed via a Probabilistic Roadmap (PRM). To introduce controlled sub-optimality, we additionally employ Greedy Best-First Search (GBFS) with a biased heuristic that encourages detours. While the standard admissible A* heuristic corresponds to the Euclidean distance to the goal in 3D space, we modify it by incorporating an attraction term toward a randomly sampled detour point $p$:
\begin{equation}
h = \beta d_g + (1 - \beta) d_p,
\label{eq:heuristic}
\end{equation}
where $d_g$ denotes the Euclidean distance to the goal, $d_p$ the Euclidean distance to the detour point $p$, and $\beta \in [0,1]$ controls the strength of the goal-directed bias. Lower values of $\beta$ increase the influence of the detour term and increase sub-optimality. 

During training, the reference path is randomly generated either using standard A* with the Euclidean heuristic or GBFS with the biased heuristic in~\eqref{eq:heuristic}. The sampling probability and the bias parameter $\beta$ shape the distribution of reference paths and therefore influence the learning dynamics. For each episode, the resulting path is post-processed using a line-of-sight smoothing procedure. Additionally, random perturbations are independently applied to each waypoint to further increase variability. As a consequence, some reference paths may become partially infeasible due to collisions or constraint violations.

\subsection{Reward Design for Opportunistic Deviation}

The navigation objective is to reach the goal region as efficiently as possible. We adopt the reward formulation of \cite{yang2025spatially} and summarize it here for completeness. The reward is purely goal-driven and contains no explicit path-following term.
At each timestep $t$, the total reward is
\begin{equation}
r_t = \alpha_1 r_t^{\text{task}} 
    + \alpha_2 r_t^{\text{reg}} 
    + \alpha_3 r_t^{\text{pen}}
    + \alpha_4 r_t^{\text{shortcut}},
\end{equation}
where $r_t^{\text{task}}$ promotes task completion, $r_t^{\text{reg}}$ enforces motion smoothness, and $r_t^{\text{pen}}$ penalizes unsafe behavior.

\noindent\textbf{Task Reward} $\mathbf{r_t^{\text{\textbf{task}}}}$ with the relative goal position $p_t$, the normalization factor $\sigma$, the episode horizon $T_{\max}$, and the terminal reward window $T_r$:
\begin{equation}
r_t^{\text{task}} 
=
\frac{
\mathbf{1}\!\left( t > T_{\max} - T_r \;\lor\; \text{random} < \delta_{\text{check}} \right)
}{
1 + \left\| \frac{p_t}{\sigma} \right\|_2
}.
\end{equation}

\noindent\textbf{Regularization} $\mathbf{r_t^{\text{\textbf{reg}}}}$ with
the executed action $a_t$, the joint accelerations  $j_t^{\text{acc}}$, and the momentum-filtered action $a_t^m = \lambda a_{t-1}^m + (1-\lambda) a_t$ with factor $\lambda$:
\begin{equation}
r_t^{\text{reg}} 
=
\beta_1 \left\| a_t - a_t^{m} \right\|_1
+
\beta_2 \left\| j_t^{\text{acc}} \right\|_1,
\end{equation}

\noindent\textbf{Penalties} $\mathbf{r_t^{\text{\textbf{pen}}}}$ with the body inclination $\theta_t$ and a safety threshold $\theta_{\text{safe}}$.
\begin{equation}
r_t^{\text{pen}}
=
\eta_1 \mathbf{1}(\text{collision})
+
\eta_2 \max\!\left(0, |\theta_t| - \theta_{\text{safe}} \right),
\end{equation}

To encourage opportunistic deviation from suboptimal or noisy reference paths, we introduce an additional shortcut reward $r_t^{\text{shortcut}}$. A shortcut is identified when the agent’s progress along the reference path exceeds a threshold of the total path length within a single simulation timestep. When this condition is met, a positive reward is granted, explicitly reinforcing behaviors that bypass unnecessary detours suggested by the reference path.
Formally, the shortcut reward is defined as:
\begin{equation}
r_t^{\text{shortcut}} = \begin{cases}
\Delta p, & \text{if } \Delta p > \epsilon \\
0,  & \text{if } \Delta p \le \epsilon
\end{cases} \label{eq:shortcut}
\end{equation}
It indicates whether a shortcut event occurred at timestep $t$, $\Delta p$ denotes the progress in percent along the path between two consecutive time steps, $\epsilon$ denotes the threshold at which we consider a shortcut, and $\alpha_4$ is a scalar weight controlling the influence of the shortcut reward. 

To summarize, the reward structure depends exclusively on goal-reaching and stability objectives; any path-following behavior must therefore emerge implicitly from the observation-conditioning and training procedure.
Finally, the policy is trained following an Asymmetric Actor-Critic PPO pipeline and with the training regularization techniques presented in \cite{yang2025spatially}, namely Temporally Consistent Dropout and Deep Mutual Learning.

\section{Experiments}

\subsection{Training and Simulation Details}
The RL training, experiments, and simulation environment are implemented in Isaac Sim using Isaac Lab~\cite{mittal2023orbit, mittal2025isaaclab} and the \texttt{rsl-rl} framework~\cite{schwarke2025rslrl}. The final policy is trained entirely in simulation for 39.5 hours on a single NVIDIA GeForce RTX 4090, with 1046 agents simulated in parallel across 180 procedurally generated environments. No dedicated curriculum is employed; instead, domain randomization is applied to the policy inputs to enhance sim-to-real generalization.
The proprioceptive observation $o_t^{prop}$ consists of the linear and angular velocities $(v_t, \omega_t \in \mathbb{R}^3)$ as well as the projected gravity vector $n_t \in \mathbb{R}^3$ and the last action $a_{t-1} \in \mathbb{R}^3$. Additionally, the reference path $\tilde{P}_t$ is composed of $N=15$ waypoints. The simulated depth camera has a horizontal field of view of $105^\circ$, a vertical field of view of $78^\circ$, and a maximum sensing range of \SI[round-precision = 0]{10}{m} with an input image $I_t \in \mathbb{R}^{40\times 64}$. Once flattened and concatenated, the policy takes as input a vector of 2636 values.
The navigation policy operates at \SI[round-precision = 0]{5}{Hz} and outputs an action $a_t$, which is executed by a separate locomotion policy running at \SI[round-precision = 0]{50}{Hz} for low-level control. The actor policy contains ~1.76M trainable parameters, from which 12,960 belong to the path encoding module.


\subsection{Experimental Setup and Evaluation Metrics}
\label{subsec:Experimental Setup and Evaluation Metrics}

After comparing the baseline policy against our final path-aware model in Section~\ref{subsec:our_vs_baseline}, we conduct a set of experiments to validate the design choices introduced in Section~\ref{sec:methodology}. 
Section~\ref{subsub:architecture_ablation} assesses the impact of architectural decisions by comparing several variants of the path encoding module against each other. 
To enable a controlled comparison, the architectural variants are trained on an optimal reference path, allowing us to isolate the model’s ability to encode and exploit path information without confounding effects from path degradation. 
Secondly, Section~\ref{subsubsec:training_ablation} evaluates different training procedures. 
In this setting, all policies are tested using degraded reference paths that are both noisy and suboptimal, in order to assess their robustness to imperfect path guidance. 

\begin{figure*}[!t]
    \centering
    \begin{subfigure}{0.28\textwidth}
        \centering
        \includegraphics[width=\linewidth]{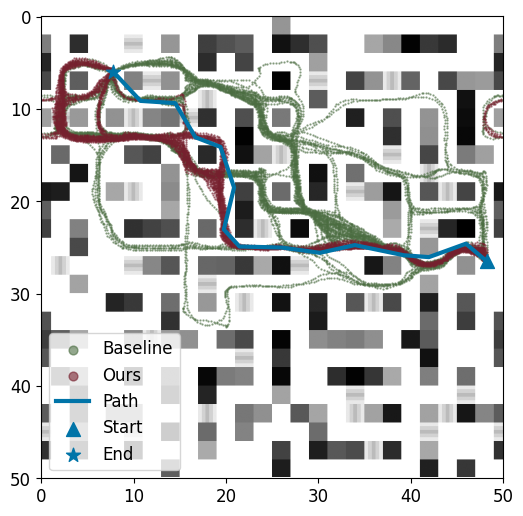}
        \caption{Scenario 1.}
        \label{fig:sub1}
    \end{subfigure}
    \begin{subfigure}{0.28\textwidth}
        \centering
        \includegraphics[width=\linewidth]{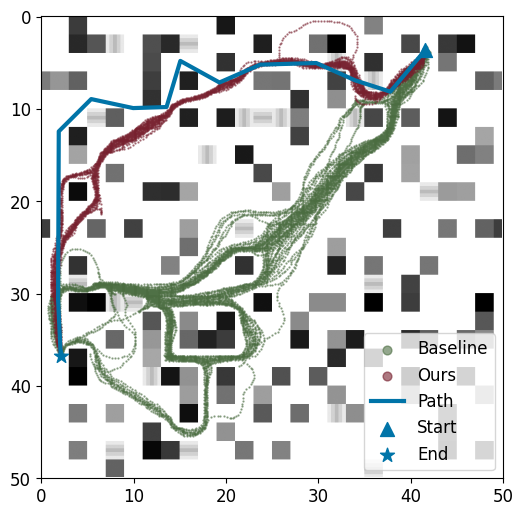}
        \caption{Scenario 2.}
        \label{fig:sub2}
    \end{subfigure}
    \begin{subfigure}{0.28\textwidth}
        \centering
        \includegraphics[width=\linewidth]{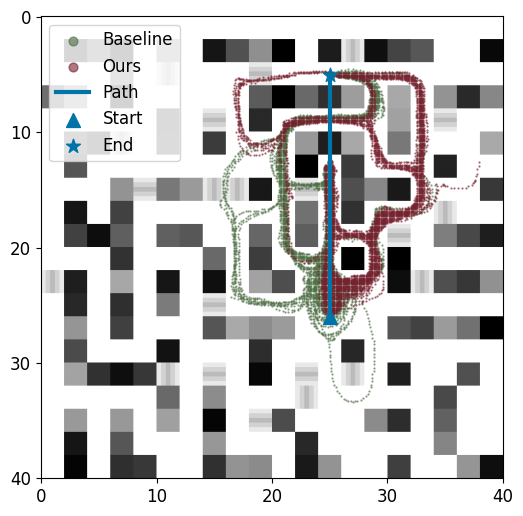}
        \caption{Scenario 3.}
        \label{fig:sub3}
    \end{subfigure}
    \caption{Qualitative comparison of navigation trajectories in a $50\times\SI[round-precision = 0]{50}{m}$ and $40\times\SI[round-precision = 0]{40}{m}$ maze environments. For each policy, 100 trajectories are shown.
    }
    \label{fig:qualitative_exp}
\end{figure*}

To perform the evaluations in Sections~\ref{subsec:our_vs_baseline}--\ref{subsubsec:training_ablation}, we establish the following evaluation protocols and metrics. We define evaluation conditions that differ from those encountered during training, in order to assess generalization. Specifically, while training is performed on environments of size $30\times\SI[round-precision = 0]{30}{m}$, evaluation is conducted on larger terrains of size $50\times\SI[round-precision = 0]{50}{m}$. The maximum episode duration is also increased from \SI[round-precision = 0]{60}{s} during training to \SI[round-precision = 0]{120}{s} at evaluation time. The procedural terrain generation process is kept identical, ensuring that obstacle distributions remain statistically consistent while increasing the spatial scale of the task. We evaluate all policies across 75 procedurally generated terrains, resulting in up to 2250 evaluation episodes (30 trajectories per terrain). When a testing episode is repeated, the starting position remains fixed while the robot’s initial orientation is randomized.
Performance is quantified primarily using the Success Rate (SR) and the Success weighted by Path Length (SPL), described in \cite{anderson2018evaluationembodiednavigationagents} and defined as
\begin{equation}
\mathrm{SPL} = \frac{1}{N} \sum_{i=1}^{N} S_i \, \frac{L_i}{\max(l_i, L_i)},
\label{eq:spl}
\end{equation}
where $L_i$ is the optimal path length, computed using the A* with the standard Euclidean distance as a heuristic, and $l_i$ is the robot's trajectory length. To provide a more fine-grained analysis, the SPL is reported as a function of the optimal path length.

We establish two evaluation reference path scenarios: (1) the agent is provided with the optimal path computed by the planner, (2) the agent observes a degraded reference path that is both noisy and suboptimal. To assess robustness beyond training conditions, waypoint noise is sampled with a displacement magnitude ranging up to \SI[round-precision = 0]{2}{m}, compared to \SI[round-precision = 0]{1}{m} in training. In addition, non-optimal paths are selected so that their total length lies between 1.25 and 3.33 times that of the corresponding optimal path. This evaluation protocol allows us to assess both the best-case performance when high-quality path information is available and the worst-case behavior when reference paths are misleading or unreliable.

\subsection{Comparison with Baseline Navigation}
\label{subsec:our_vs_baseline}

We compare our path-aware model against the baseline policy from \cite{yang2025spatially} under the two reference path observation scenarios introduced in \ref{subsec:Experimental Setup and Evaluation Metrics}. Table~\ref{tab:model_performance} summarizes the SR and SPL results for both policies. When provided with an optimal reference path, the path-aware model achieves a 7.02\% increase in SPL compared to the baseline, demonstrating its ability to effectively exploit high-quality path information. Under degraded path observations, the performance of our model remains comparable to the baseline, indicating that misleading or unreliable path information does not degrade navigation performance. These results highlight that the proposed method improves navigation efficiency when informative path observations are available, while maintaining baseline-level performance when path information is poor or unreliable.

\begin{table}[b]
\centering
\captionsetup{width=0.88\linewidth}
\caption{Performance comparison between the baseline and our model over 2250 episodes for each.}
\label{tab:model_performance}
\setlength{\tabcolsep}{3pt}
\begin{tabular}{lSS}
\hline
\textbf{Model} & {SR} & {SPL} \\
\hline
Baseline & 0.8320 & 0.7463 \\
Our model (optimal path) & \bfseries 0.8658 & \bfseries 0.8165 \\
Our model (non-optimal \& noisy) & 0.8276 & 0.7427 \\
\hline
\end{tabular}
\end{table}

We also conduct qualitative experiments to analyze the behavior of the proposed path-aware policy compared to the baseline. As illustrated in Fig.~\ref{fig:qualitative_exp}, the path-aware model consistently exploits the provided path information to guide its exploration. Compared to the baseline, the resulting trajectories exhibit significantly lower variability and greater consistency across episodes. Even when the suggested path is suboptimal (b), the path-aware agent generally benefits from the high-level directional cues, leading to more efficient navigation and fewer unnecessary detours. Furthermore, it can still reach the goal when the reference path is blocked, and additional exploration is required (c). These qualitative results, together with the results from Table~\ref{tab:model_performance}, suggest that the proposed approach successfully induces the desired behavior, guiding exploration while preserving the ability to deviate from imperfect guidance when beneficial.

\subsection{Ablation Study}
\label{subsec:ablation}
We now analyze which design choices contribute to this improvement.
\subsubsection{Model Architecture Ablation}
\label{subsub:architecture_ablation}

We first evaluate the capacity of the proposed architecture to exploit reference path information. To isolate this effect, we consider an idealized setting in which the observed path corresponds to the optimal trajectory computed by the global planner. This experiment is designed to assess whether the policy can leverage path information purely through its representation and integration within the network. We compare several variants of the path encoding module, differing in how the path representation is processed before being concatenated with the rest of the policy features:
\begin{enumerate}
    \item direct concatenation of the raw path representation,
    \item only self-attention-based path encoding,
    \item self- and cross-attention with a fixed random query,
    \item self- and cross-attention with a learned query.
\end{enumerate}

\begin{figure}[!t]
    \centering
    \includegraphics[width=.9\linewidth]{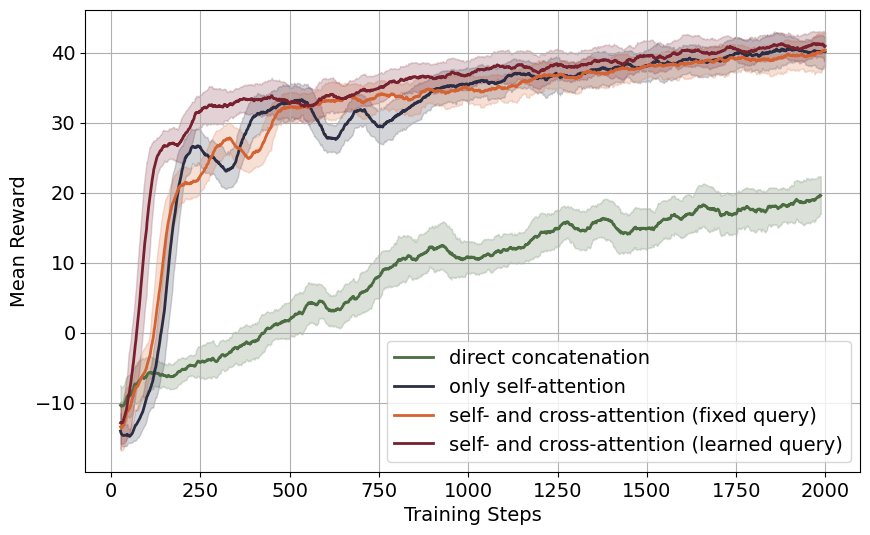}
    \caption{Comparison of training reward across different path encoding architectures, illustrating the effect of architectural choices on learning performance.}
    \label{fig:path follower comparision}
\end{figure}

As shown in Fig.~\ref{fig:path follower comparision}, the learned-query variant exhibits the steepest initial slope of the learning curve, indicating the fastest rate of performance improvement during early training. It also maintains consistently higher rewards than the other variants over the first 2000 training iterations. We therefore adopted this variant for our final model and all subsequent experiments.


\subsubsection{Training Process Ablation}
\label{subsubsec:training_ablation}

We then evaluate the robustness of path-aware policy variants obtained from different training procedures using the SR and SPL metric with the considered policies specified in Table~\ref{tab:model_components}.
All models are evaluated under the poor path quality scenario defined in \ref{subsec:Experimental Setup and Evaluation Metrics}. 
Table~\ref{tab:spl_sr_by_distance} summarizes the SR and SPL results across all ablations.

\begin{table}[b]
\centering
\captionsetup{width=0.88\linewidth}
\caption{Training configuration and ablation settings.}
\label{tab:model_components}
\begin{tabular}{lccc}
\hline
\textbf{Model} & \textbf{\shortstack{Sub-optimal\\Path}} & \textbf{\shortstack{Path\\Noise}} & \textbf{\shortstack{Shortcut\\Reward}} \\
\hline
Path Follower & no & no & no \\
\hline
\shortstack{Reward \& Noise \\ Ablation} & yes & no & no \\
\hline
Reward Ablation & yes & yes &no \\
\hline
Noise Ablation & yes & no & yes \\
\hline
Full Model & yes & yes & yes \\
\hline
\end{tabular}
\end{table}

\begin{table*}[!t]
\centering
\caption{SR and SPL as a function of optimal path length.}
\label{tab:spl_sr_by_distance}
\setlength{\tabcolsep}{3pt}
\begin{tabular}{l*{14}{S}}
\hline
& \multicolumn{2}{c}{0--10m} 
& \multicolumn{2}{c}{10--20m} 
& \multicolumn{2}{c}{20--30m} 
& \multicolumn{2}{c}{30--40m} 
& \multicolumn{2}{c}{40--50m} 
& \multicolumn{2}{c}{50--60m} 
& \multicolumn{2}{c}{60--70m} \\
Model 
& SPL & SR 
& SPL & SR 
& SPL & SR 
& SPL & SR 
& SPL & SR 
& SPL & SR 
& SPL & SR \\
\hline
Path follower 
& 0.3923 & 0.6333 
& 0.2358 & 0.3973 
& 0.1932 & 0.3217 
& 0.1472 & 0.2472 
& 0.1261 & 0.1918 
& 0.0677 & 0.0909 
& 0.0410 & 0.0455 \\

Reward \& Noise Ablation 
& 0.4251 & 0.5393 
& 0.3769 & 0.4548 
& 0.3413 & 0.3915 
& 0.3474 & 0.4075 
& 0.3825 & 0.4503 
& 0.3051 & 0.3788 
& 0.2576 & 0.2778 \\

Full Model 
& 0.7923 & 0.8947 
& 0.8158 & 0.8949 
& 0.8040 & \bfseries 0.8786
& 0.7431 & \bfseries 0.8243
& 0.7130 & 0.8201 
& 0.6168 & \bfseries 0.7125 
& \bfseries 0.6295 & \bfseries 0.6957 \\

Reward Ablation 
& \bfseries 0.8313 & \bfseries 0.9231
& \bfseries 0.8378 & \bfseries 0.9050
& \bfseries 0.8100 & 0.8706 
& \bfseries 0.7472 & 0.8184 
& \bfseries 0.7243 & \bfseries 0.8208
& \bfseries 0.6217 & 0.6905 
& 0.5046 & 0.5417 \\

Noise Ablation 
& 0.2238 & 0.2941 
& 0.2088 & 0.2485 
& 0.1935 & 0.2327 
& 0.2085 & 0.2423 
& 0.1817 & 0.2203 
& 0.1217 & 0.1429 
& 0.1544 & 0.1818 \\
\hline
\end{tabular}
\end{table*}

The results indicate that models trained without waypoint noise exhibit significantly reduced robustness when evaluated on degraded path observations. While all path-aware variants outperform rigid path-following behavior, the full model achieves the highest average SR. Although the overall performance gap between the full model and the shortcut-reward ablated variant is modest, the full model exhibits significantly less degradation on longer trajectories.

\subsection{Robustness to Missing Inputs}
\label{subsec:missing_input}
\begin{figure}[t]
    \centering
    \includegraphics[width=0.9\linewidth]{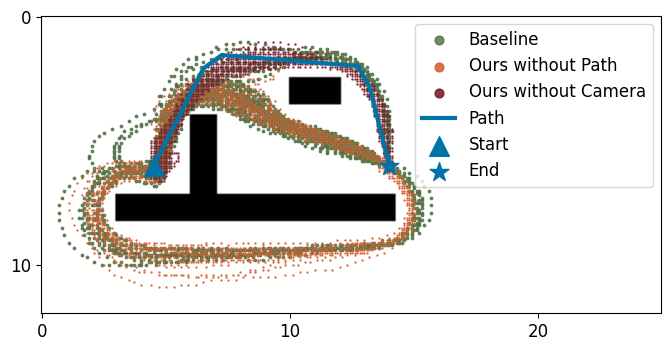}
    \caption{Qualitative comparison of a 100 navigation trajectories under input ablation, where the scale is in meters. 
    }
    \label{fig:input_ablation}
\end{figure}

We evaluate the robustness of the policy to missing inputs by simulating extreme sensing conditions in which specific input modalities are replaced with zero tensors, as illustrated in Fig.~\ref{fig:input_ablation}. Two scenarios are considered: (1) removing the reference path observation and (2) removing the depth image input. When the reference path is ablated, the agent behaves similarly to the baseline, reaching the goal through unguided exploration even in larger $50\times\SI[round-precision = 0]{50}{m}$ environments. This indicates that the policy can rely solely on perceptual inputs when path information is unavailable. Conversely, when the depth input is removed, the agent exhibits path-following behavior, navigating primarily based on the waypoint sequence. Although successful execution is mainly observed for shorter, less perturbed paths, this demonstrates that the model can effectively interpret and exploit structured path information in the absence of visual input.

To further assess the robustness and transferability of our approach, we deployed the trained policy on a Unitree B2W quadruped in a previously unseen indoor environment. The robot is equipped with a ZEDX depth camera and an NVIDIA Jetson AGX Orin for onboard computation. Proprioceptive observations $o_t^{prop}$ are provided by a LiDAR-based state estimation system. 

\begin{figure*}[t]
    \centering
    \includegraphics[width=.95\textwidth]{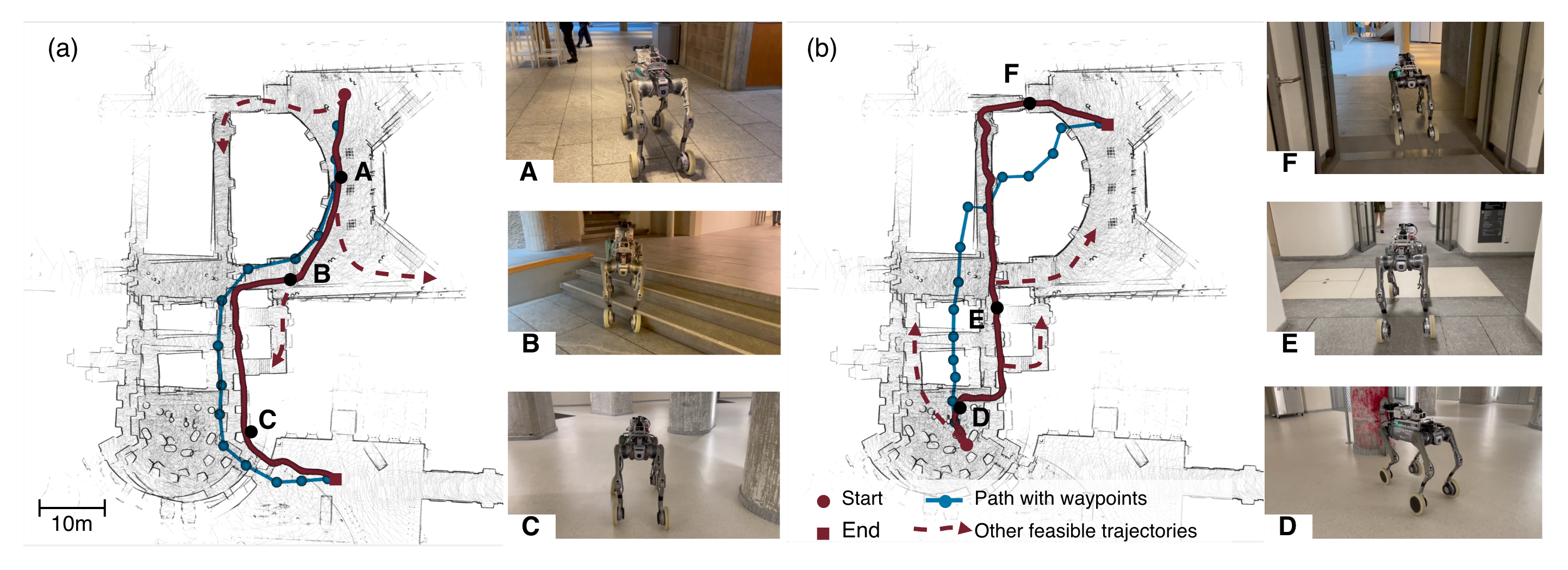}
    \caption{Real-world deployment in a university building with long-range navigation trajectories. The robot trajectory shown by the continuous red line spans about \SI[round-precision = 0]{93}{m} in scenario (a) and \SI[round-precision = 0]{91}{m} in scenario (b).}
    \label{fig:real_world_test}
\end{figure*}

As illustrated in Fig.~\ref{fig:real_world_test}, the real-world results are consistent with the behaviors observed in simulation, particularly in long-range navigation scenarios involving extended traversals and multiple directional commitments. Importantly, the reference path is intentionally selected to reduce stair traversal rather than to minimize distance. Even when the provided path is imperfect, the policy leverages it to bias exploration and avoids entering corridors that would yield trajectories of comparable length but increased uncertainty. More specifically, the dashed red arrows indicate alternative routes that would have reached the goal with similar path length but required negotiating more stairs, resulting in less reliable and less safe terrain for the locomotion policy; some alternatives also lead to dead ends or longer detours. Together, these experiments demonstrate successful sim-to-real transfer and highlight the model’s capacity to integrate global path information with local perception, biasing the robot toward safer and more desirable trajectories while preserving goal-reaching performance.

\section{Discussion}

\subsection{Opportunistic Use of Path Information}

We observe that training with exclusively optimal reference paths promotes rigid path-following behavior, whereas exposure to imperfect or infeasible paths encourages the policy to learn when and how to trust path information. Consistent with this, the experimental results indicate that the proposed model exploits path guidance opportunistically rather than blindly following it. When the reference path is near-optimal and reliable, the policy improves navigation efficiency by reducing unnecessary exploration, dead ends, and detours. Conversely, when the path is misleading or suboptimal, the agent prioritizes goal-directed reasoning over strict adherence to the provided guidance.


This behavior can be attributed to the combination of implicit path conditioning and the underlying reward structure. Although the model is trained on suboptimal, noisy paths, the reward function remains focused on reaching the goal efficiently, without explicitly rewarding path-following. As a result, the policy is encouraged to treat the path as contextual guidance rather than as an execution constraint. Implicit conditioning, therefore, enables the model to modulate its behavior based on the observed path while preserving its fundamental objective of reaching the goal as efficiently as possible.





\subsection{Fixed Path Representation}

The current model represents the reference path using a fixed number of waypoints, which limits the number of turns and the spatial extent it can represent. While this choice is sufficient for the environments' size considered in this work, it may restrict scalability to longer-range navigation scenarios spanning hundreds of meters. Future work could investigate the impact of increasing the waypoint count or, more generally, develop architectures capable of processing paths with a variable number of waypoints.

\subsection{Implicit Modeling of Path Reliability}

The policy implicitly learns how much to rely on the reference path through exposure to suboptimal and noisy paths during training. While this enables opportunistic use of path information, it also introduces two limitations. First, the resulting behavior depends strongly on the training path distribution, making it difficult to precisely tune or predict the learned reliance on path information. Second, at inference time, the policy treats all observed paths uniformly and cannot explicitly distinguish between accurate and unreliable guidance.

A natural extension would be to incorporate explicit reasoning about path quality. Path reliability could be provided as an additional observation by a high-level planner or estimated online by the policy itself. For instance, path features could be gated by the SRU's internal state, which encodes the agent’s recent experience and interaction history. Such a mechanism would allow the policy to dynamically modulate its reliance on the reference path.

\section{Conclusion}

We proposed a path-aware local navigation policy that leverages reference path information opportunistically rather than treating it as an explicit execution constraint. By conditioning an RL policy on imperfect and potentially misleading path observations, our approach addresses key limitations of purely exploratory, goal-oriented navigation, such as unnecessary detours and dead-end exploration.

Our experimental results demonstrate that access to path information significantly improves navigation efficiency when high-quality paths are available, while performance gracefully degrades to baseline levels when path observations are severely degraded. This behavior highlights the robustness of implicit path conditioning and distinguishes it from rigid path-following approaches. The proposed method is particularly well-suited for long-range navigation scenarios in which high-level planning is inaccurate, but local execution must remain adaptive to uncertainty and environmental variability.

\ifthenelse{\boolean{anonymous}}%
{} 
{ 
\section*{ACKNOWLEDGMENT}
This research was supported by the Swiss National Science Foundation (SNSF) as part of project No.227617.
}

\bibliographystyle{IEEEtran}
\bibliography{references}

@IEEEtranBSTCTL{BSTcontrol,
  CTLuse_forced_etal       = "yes",
  CTLmax_names_forced_etal = "10",
  CTLnames_show_etal       = "11",
  CTLdash_repeated_names   = "no"
}

@article{yang2025spatially,
  title={Spatially-Enhanced Recurrent Memory for Long-Range Mapless Navigation via End-to-End Reinforcement Learning},
  author={Yang, Fan and Frivik, Per and Hoeller, David and Wang, Chen and Cadena, Cesar and Hutter, Marco},
  journal={International Journal of Robotics Research},
  year={2025},
  archivePrefix={arXiv},
  eprint={2506.05997},
  primaryClass={cs.RO}
}

@article{shi2019end,
  title={End-to-end navigation strategy with deep reinforcement learning for mobile robots},
  author={Shi, Haobin and Shi, Lin and Xu, Meng and Hwang, Kao-Shing},
  journal={IEEE Transactions on Industrial Informatics},
  volume={16},
  number={4},
  pages={2393--2402},
  year={2019},
  publisher={IEEE}
}

@article{cimurs2021goal,
  title={Goal-driven autonomous exploration through deep reinforcement learning},
  author={Cimurs, Reinis and Suh, Il Hong and Lee, Jin Han},
  journal={IEEE Robotics and Automation Letters},
  volume={7},
  number={2},
  pages={730--737},
  year={2021},
  publisher={IEEE}
}

@article{schubert2021plan,
  title={Plan-based relaxed reward shaping for goal-directed tasks},
  author={Schubert, Ingmar and Oguz, Ozgur S and Toussaint, Marc},
  journal={arXiv preprint arXiv:2107.06661},
  year={2021}
}

@article{merton2024deep,
  title={Deep Reinforcement Learning for Local Path Following of an Autonomous Formula SAE Vehicle},
  author={Merton, Harvey and Delamore, Thomas and Stol, Karl and Williams, Henry},
  journal={arXiv preprint arXiv:2401.02903},
  year={2024}
}

@article{schubert2021learning,
  title={Learning to execute: Efficient learning of universal plan-conditioned policies in robotics},
  author={Schubert, Ingmar and Driess, Danny and Oguz, Ozgur S and Toussaint, Marc},
  journal={Advances in Neural Information Processing Systems},
  volume={34},
  pages={1912--1924},
  year={2021}
}

@article{10.1145/3727642,
author = {Nahavandi, Saeid and Alizadehsani, Roohallah and Nahavandi, Darius and Mohamed, Shady and Mohajer, Navid and Rokonuzzaman, Mohammad and Hossain, Ibrahim},
title = {A Comprehensive Review on Autonomous Navigation},
year = {2025},
issue_date = {September 2025},
publisher = {Association for Computing Machinery},
address = {New York, NY, USA},
volume = {57},
number = {9},
issn = {0360-0300},
url = {https://doi.org/10.1145/3727642},
doi = {10.1145/3727642},
journal = {ACM Comput. Surv.},
month = may,
articleno = {234},
numpages = {67}
}

@inproceedings{shah2022viking, 
    author    = {Dhruv Shah and Sergey Levine}, 
    title     = {{ViKiNG: Vision-Based Kilometer-Scale Navigation with Geographic Hints}}, 
    booktitle = {Proceedings of Robotics: Science and Systems}, 
    year      = {2022},
    url      = {http://www.roboticsproceedings.org/rss18/p019.html} 
}

@article{fan2018crowdmove,
  title={Crowdmove: Autonomous mapless navigation in crowded scenarios},
  author={Fan, Tingxiang and Cheng, Xinjing and Pan, Jia and Manocha, Dinesh and Yang, Ruigang},
  journal={arXiv preprint arXiv:1807.07870},
  year={2018}
}

@article{Lee_2024,
   title={Learning robust autonomous navigation and locomotion for wheeled-legged robots},
   volume={9},
   ISSN={2470-9476},
   url={http://dx.doi.org/10.1126/scirobotics.adi9641},
   DOI={10.1126/scirobotics.adi9641},
   number={89},
   journal={Science Robotics},
   publisher={American Association for the Advancement of Science (AAAS)},
   author={Lee, Joonho and Bjelonic, Marko and Reske, Alexander and Wellhausen, Lorenz and Miki, Takahiro and Hutter, Marco},
   year={2024},
   month=apr }

@ARTICLE{9205217,
  author={Wang, Binyu and Liu, Zhe and Li, Qingbiao and Prorok, Amanda},
  journal={IEEE Robotics and Automation Letters}, 
  title={Mobile Robot Path Planning in Dynamic Environments Through Globally Guided Reinforcement Learning}, 
  year={2020},
  volume={5},
  number={4},
  pages={6932-6939},
  doi={10.1109/LRA.2020.3026638}}

@misc{nguyen2025emergencedeepreinforcementlearning,
      title={The Emergence of Deep Reinforcement Learning for Path Planning}, 
      author={Thanh Thi Nguyen and Saeid Nahavandi and Imran Razzak and Dung Nguyen and Nhat Truong Pham and Quoc Viet Hung Nguyen},
      year={2025},
      eprint={2507.15469},
      archivePrefix={arXiv},
      primaryClass={cs.RO},
      url={https://arxiv.org/abs/2507.15469}, 
}

@Article{app12146874,
AUTHOR = {Cheng, Xiuquan and Zhang, Shaobo and Cheng, Sizhu and Xia, Qinxiang and Zhang, Junhao},
TITLE = {Path-Following and Obstacle Avoidance Control of Nonholonomic Wheeled Mobile Robot Based on Deep Reinforcement Learning},
JOURNAL = {Applied Sciences},
VOLUME = {12},
YEAR = {2022},
NUMBER = {14},
ARTICLE-NUMBER = {6874},
URL = {https://www.mdpi.com/2076-3417/12/14/6874},
ISSN = {2076-3417},
DOI = {10.3390/app12146874}
}

@article{mittal2023orbit,
  title={Orbit: A Unified Simulation Framework for Interactive Robot Learning Environments}, 
  author={Mittal, Mayank and Yu, Calvin and Yu, Qinxi and Liu, Jingzhou and Rudin, Nikita and Hoeller, David and Yuan, Jia Lin and Singh, Ritvik and Guo, Yunrong and Mazhar, Hammad and Mandlekar, Ajay and Babich, Buck and State, Gavriel and Hutter, Marco and Garg, Animesh},
  journal={IEEE Robotics and Automation Letters}, 
  year={2023},
  volume={8},
  number={6},
  pages={3740-3747},
  doi={10.1109/LRA.2023.3270034}
}

@article{mittal2025isaaclab,
  title={Isaac Lab: A GPU-Accelerated Simulation Framework for Multi-Modal Robot Learning},
  author={Mayank Mittal and Pascal Roth and James Tigue and Antoine Richard and Octi Zhang and Peter Du and Antonio Serrano-Muñoz and Xinjie Yao and René Zurbrügg and Nikita Rudin and Lukasz Wawrzyniak and Milad Rakhsha and Alain Denzler and Eric Heiden and Ales Borovicka and Ossama Ahmed and Iretiayo Akinola and Abrar Anwar and Mark T. Carlson and Ji Yuan Feng and Animesh Garg and Renato Gasoto and Lionel Gulich and Yijie Guo and M. Gussert and Alex Hansen and Mihir Kulkarni and Chenran Li and Wei Liu and Viktor Makoviychuk and Grzegorz Malczyk and Hammad Mazhar and Masoud Moghani and Adithyavairavan Murali and Michael Noseworthy and Alexander Poddubny and Nathan Ratliff and Welf Rehberg and Clemens Schwarke and Ritvik Singh and James Latham Smith and Bingjie Tang and Ruchik Thaker and Matthew Trepte and Karl Van Wyk and Fangzhou Yu and Alex Millane and Vikram Ramasamy and Remo Steiner and Sangeeta Subramanian and Clemens Volk and CY Chen and Neel Jawale and Ashwin Varghese Kuruttukulam and Michael A. Lin and Ajay Mandlekar and Karsten Patzwaldt and John Welsh and Huihua Zhao and Fatima Anes and Jean-Francois Lafleche and Nicolas Moënne-Loccoz and Soowan Park and Rob Stepinski and Dirk Van Gelder and Chris Amevor and Jan Carius and Jumyung Chang and Anka He Chen and Pablo de Heras Ciechomski and Gilles Daviet and Mohammad Mohajerani and Julia von Muralt and Viktor Reutskyy and Michael Sauter and Simon Schirm and Eric L. Shi and Pierre Terdiman and Kenny Vilella and Tobias Widmer and Gordon Yeoman and Tiffany Chen and Sergey Grizan and Cathy Li and Lotus Li and Connor Smith and Rafael Wiltz and Kostas Alexis and Yan Chang and David Chu and Linxi "Jim" Fan and Farbod Farshidian and Ankur Handa and Spencer Huang and Marco Hutter and Yashraj Narang and Soha Pouya and Shiwei Sheng and Yuke Zhu and Miles Macklin and Adam Moravanszky and Philipp Reist and Yunrong Guo and David Hoeller and Gavriel State},
  journal={arXiv preprint arXiv:2511.04831},
  year={2025},
  url={https://arxiv.org/abs/2511.04831}
}

@article{schwarke2025rslrl,
  title={RSL-RL: A Learning Library for Robotics Research},
  author={Schwarke, Clemens and Mittal, Mayank and Rudin, Nikita and Hoeller, David and Hutter, Marco},
  journal={arXiv preprint arXiv:2509.10771},
  year={2025}
}

@misc{anderson2018evaluationembodiednavigationagents,
      title={On Evaluation of Embodied Navigation Agents}, 
      author={Peter Anderson and Angel Chang and Devendra Singh Chaplot and Alexey Dosovitskiy and Saurabh Gupta and Vladlen Koltun and Jana Kosecka and Jitendra Malik and Roozbeh Mottaghi and Manolis Savva and Amir R. Zamir},
      year={2018},
      eprint={1807.06757},
      archivePrefix={arXiv},
      primaryClass={cs.AI},
      url={https://arxiv.org/abs/1807.06757}, 
}

@article{Hart1968,
  doi = {10.1109/tssc.1968.300136},
  url = {https://doi.org/10.1109/tssc.1968.300136},
  year = {1968},
  publisher = {Institute of Electrical and Electronics Engineers ({IEEE})},
  volume = {4},
  number = {2},
  pages = {100--107},
  author = {Peter Hart and Nils Nilsson and Bertram Raphael},
  title = {A Formal Basis for the Heuristic Determination of Minimum Cost Paths},
  journal = {{IEEE} Transactions on Systems Science and Cybernetics}
}

@book{alma9936230413403606,
year = {2004},
publisher = {MIT Press},
author = {Siegwart, Roland. and Nourbakhsh, Illah Reza},
address = {Cambridge, Mass. ;},
booktitle = {Introduction to autonomous mobile robots},
series = {Intelligent robots and autonomous agents},
isbn = {026219502X},
language = {eng},
lccn = {2003059349},
title = {Introduction to autonomous mobile robots / Roland Siegwart and Illah R. Nourbakhsh.},
}

@article{richter2026large,
  title={Large-Scale Autonomous Gas Monitoring for Volcanic Environments: A Legged Robot on Mount Etna},
  author={Richter, Julia and Tuna, Turcan and Patel, Manthan and Miki, Takahiro and Higgins, Devon and Fox, James and Cadena, Cesar and Diaz, Andres and Hutter, Marco},
  journal={arXiv preprint arXiv:2601.07362},
  year={2026}
}

@article{mattamala2024autonomous,
  title={Autonomous forest inventory with legged robots: system design and field deployment},
  author={Mattamala, Mat{\'\i}as and Chebrolu, Nived and Casseau, Benoit and Frei{\ss}muth, Leonard and Frey, Jonas and Tuna, Turcan and Hutter, Marco and Fallon, Maurice},
  journal={arXiv preprint arXiv:2404.14157},
  year={2024}
}

@inproceedings{hentschel2010autonomous,
  title={Autonomous robot navigation based on openstreetmap geodata},
  author={Hentschel, Matthias and Wagner, Bernardo},
  booktitle={13th International IEEE Conference on Intelligent Transportation Systems},
  pages={1645--1650},
  year={2010},
  organization={IEEE}
}

@article{truong2024indoorsim,
  title={Indoorsim-to-outdoorreal: Learning to navigate outdoors without any outdoor experience},
  author={Truong, Joanne and Zitkovich, April and Chernova, Sonia and Batra, Dhruv and Zhang, Tingnan and Tan, Jie and Yu, Wenhao},
  journal={IEEE Robotics and Automation Letters},
  volume={9},
  number={5},
  pages={4798--4805},
  year={2024},
  publisher={IEEE}
}

@article{kavraki2002probabilistic,
  title={Probabilistic roadmaps for path planning in high-dimensional configuration spaces},
  author={Kavraki, Lydia E and Svestka, Petr and Latombe, J-C and Overmars, Mark H},
  journal={IEEE transactions on Robotics and Automation},
  volume={12},
  number={4},
  pages={566--580},
  year={2002},
  publisher={IEEE}
}

@article{LaValle1998RapidlyexploringRT,
  title={Rapidly-exploring random trees : a new tool for path planning},
  author={Steven M. LaValle},
  journal={The annual research report},
  year={1998},
  url={https://api.semanticscholar.org/CorpusID:14744621}
}

@article{zhang2026focusnav,
  title={FocusNav: Spatial Selective Attention with Waypoint Guidance for Humanoid Local Navigation},
  author={Zhang, Yang and Ma, Jianming and Yan, Liyun and Cao, Zhanxiang and Zhang, Yazhou and Li, Haoyang and Gao, Yue},
  journal={arXiv preprint arXiv:2601.12790},
  year={2026}
}

@article{shah2023vint,
  title={ViNT: A foundation model for visual navigation},
  author={Shah, Dhruv and Sridhar, Ajay and Dashora, Nitish and Stachowicz, Kyle and Black, Kevin and Hirose, Noriaki and Levine, Sergey},
  journal={arXiv preprint arXiv:2306.14846},
  year={2023}
}

@inproceedings{pokle2019deep,
  title={Deep local trajectory replanning and control for robot navigation},
  author={Pokle, Ashwini and Mart{\'\i}n-Mart{\'\i}n, Roberto and Goebel, Patrick and Chow, Vincent and Ewald, Hans M and Yang, Junwei and Wang, Zhenkai and Sadeghian, Amir and Sadigh, Dorsa and Savarese, Silvio and others},
  booktitle={2019 international conference on robotics and automation (ICRA)},
  pages={5815--5822},
  year={2019},
  organization={IEEE}
}
\end{document}